\documentclass{article}

\PassOptionsToPackage{numbers}{natbib}
\usepackage[final]{nips_2016}

\usepackage[utf8]{inputenc} 
\usepackage[T1]{fontenc}    
\usepackage{hyperref}       
\usepackage{url}            
\usepackage{booktabs}       
\usepackage{amsfonts}       
\usepackage{mathtools}
\usepackage{nicefrac}       
\usepackage{microtype}      
\usepackage{graphicx}
\usepackage{caption}
\usepackage{subcaption}
\usepackage{float}
\usepackage{enumitem}
\usepackage{amsmath}
\usepackage{braket}
\usepackage{tikz}
\usepackage{afterpage}
\usetikzlibrary{arrows, chains, fit, shapes, decorations.pathreplacing, decorations.pathmorphing, positioning, calc, hobby}
\usetikzlibrary{shapes,arrows} 

\newcount\colvecbcount
\newcommand*\colvecb[1]{
    \global\colvecbcount#1
    \begin{bmatrix}
    \colvecbnext
}
\def\colvecbnext#1{
    #1
    \global\advance\colvecbcount-1
    \ifnum\colvecbcount>0
            \\
            \expandafter\colvecbnext
    \else
            \end{bmatrix}
    \fi
}

\title{CycleGAN with Better Cycles}

\author{
  Tongzhou Wang\\
  The Department of Electrical Engineering and Computer Sciences\\
  University of California, Berkeley\\
  Berkeley, CA, 94704 \\
  \texttt{simon.l@berkeley.edu} \\
  24438425
  \And
  Yihan Lin\\
  The Department of Electrical Engineering and Computer Sciences\\
  University of California, Berkeley\\
  Berkeley, CA, 94704 \\
  \texttt{linyh@berkeley.edu} \\
  3032523196
}

\begin{document}

\maketitle
\suppressfloats

\begin{abstract}
  CycleGAN provides a framework to train image-to-image translation with unpaired datasets using cycle consistency loss \cite{zhu2017unpaired}. While results are great in many applications, the pixel level cycle consistency can potentially be problematic and causes unrealistic images in certain cases. In this project, we propose three simple modifications to cycle consistency, and show that such an approach achieves better results with less artifacts.
\end{abstract}


\section{Introduction}
Image-to-image translation generates some of the most fascinating and exciting results in computer vision. Using generative adversarial networks (GANs), pix2pix gained a huge amount of popularity on Twitter with its edges to cats translation \cite{isola2016image}. Trained using unpaired data, Cycle-Generative Adversarial Networks (CycleGAN) achieves amazing translation results in many cases where paired data is impossible, such as Monet paintings to photos, zebras to horses, etc. Image-to-image translation is also important in the task of domain adaptation. For safety reasons, robots are often trained in simulated environment and using synthesized data. In order for such trained robots to behave well in real life scenarios, one possible approach is to translate real life data, e.g., images, into data similar to what they are trained with using image-to-image translation techniques.

In this paper, we identify some existing problems with the CycleGAN framework specifically with respect to the cycle consistency loss, and several modifications aiming to solve the issues.




\section{CycleGAN}

\afterpage{%
  \begin{figure}
    \centering
    \includegraphics[scale = 0.3]{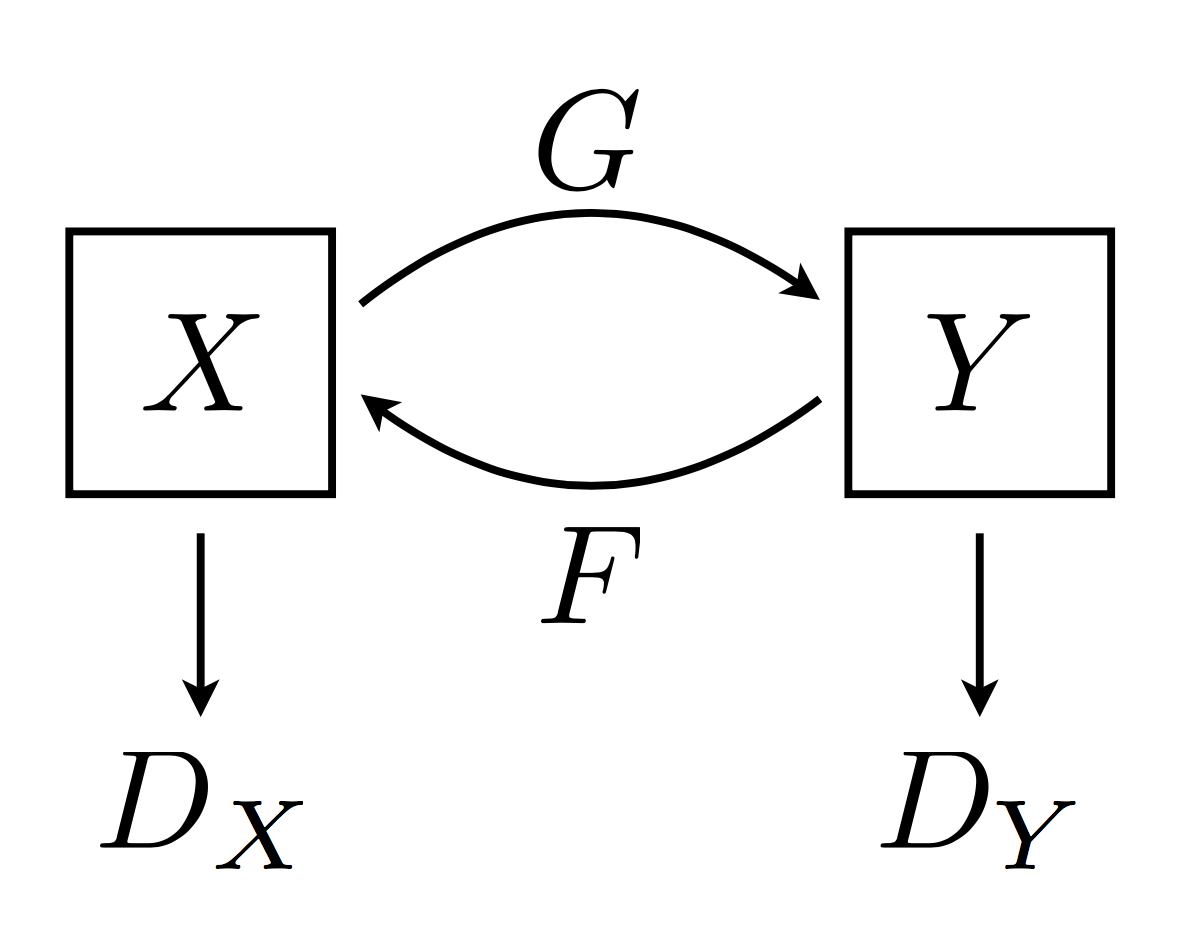}
    \caption{CycleGAN architecture.\protect\footnotemark}
    \label{fig:cycle-gan}
  \end{figure}
  \footnotetext{Figure adapted from \cite{zhu2017unpaired}.}
}

\afterpage{%
  \begin{figure}
    \centering
    \includegraphics[scale = 0.3]{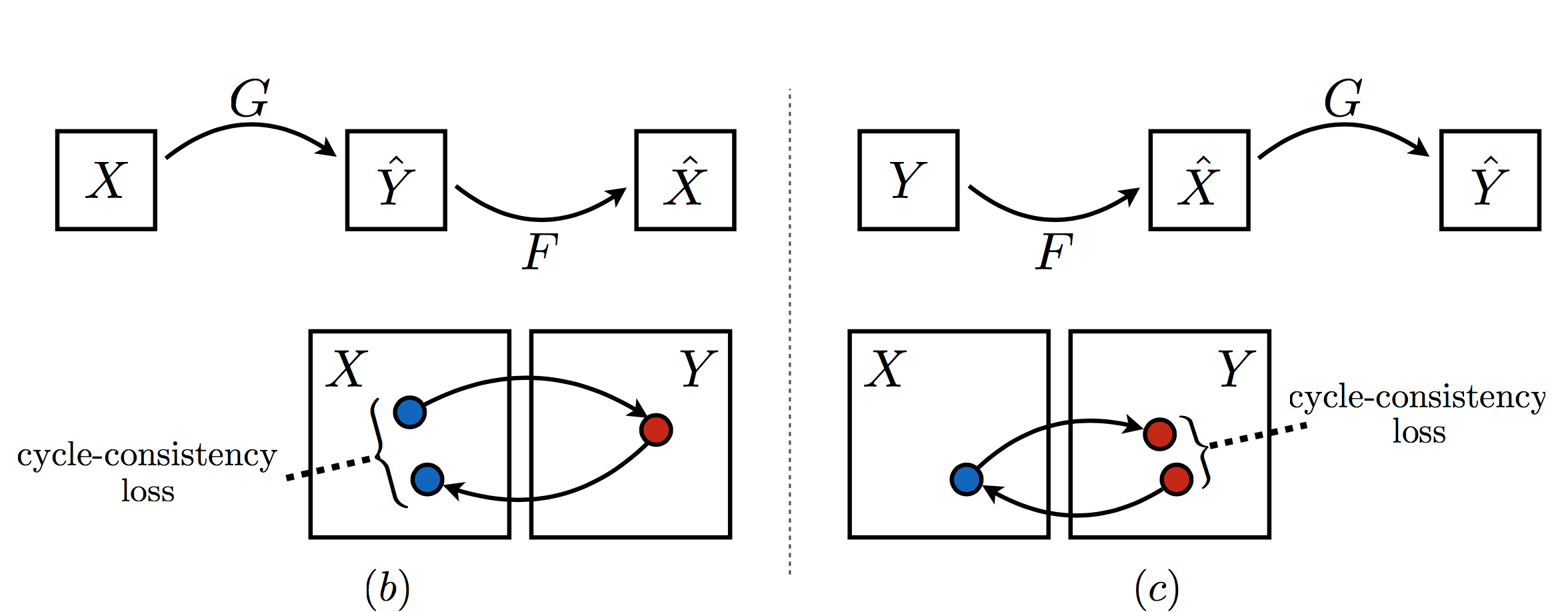}
    \caption{Cycle consistency.\protect\footnotemark}
    \label{fig:cycle-consist}
  \end{figure}
  \footnotetext{Figure adapted from \cite{zhu2017unpaired}.}
}

CycleGAN is a framework that learns image-to-image translation from unpaired datasets \cite{zhu2017unpaired}. Its architecture contains two generators and two discriminators as shown in Figure~\ref{fig:cycle-gan}. The two image domains of interest are denoted as $X$ and $Y$. Generator $G$ takes an image from $X$ as input and tries to generate a realistic image in $Y$ that tricks discriminator $D_X$. Similarly, generator $F$ generates image in reverse direction and tries to trick discriminator $D_Y$.

Similar to usual GAN settings, the discriminators encourage generators to output realistic images using the GAN loss:
\begin{equation}
  \mathcal{L}_\text{GAN}(G, D_Y, X, Y) = \mathbb{E}_{y \sim p_\text{data}(y)}[\log D_Y(y)] + \mathbb{E}_{x \sim p_\text{data}(x)}[\log (1 - D_Y(G(x)))]
\end{equation}

In order to train with unpaired data, CycleGAN proposes the notion of cycle consistency. It asserts that given a real image $x$ in $X$, if the two generators $G$ and $F$ are good, mapping it to domain $Y$ and then back to $X$ should give back the original image $x$, i.e., $x \rightarrow G(x) \rightarrow F(G(x)) \approx x$. Similarly, the backward direction should also have $y \rightarrow F(y) \rightarrow G(F(y)) \approx y$. Figure~\ref{fig:cycle-consist} graphically shows the idea of cycle consistency, which is enforced through the following cycle consistency loss:

\begin{equation}
  \mathcal{L}_\text{cyc}(G, F, X) = \mathbb{E}_{x \sim p_\text{data}(x)}[\lVert F(G(x)) - x \rVert_1]
\end{equation}

Putting the two losses together, the full objective for CycleGAN is:

\begin{align}
  \mathcal{L}(G, F, D_X, D_Y) & = \mathcal{L}_\text{GAN}(G, D_Y, X, Y) + \mathcal{L}_\text{GAN}(F, D_X, Y, X) \\
  & + \lambda \mathcal{L}_\text{cyc}(G, F, X) + \lambda \mathcal{L}_\text{cyc}(F, G, Y)
\end{align}

\subsection{Effects of cycle consistency}
\label{sec:cycle}
On a high level, cycle consistency encourages generators to avoid unnecessary changes and thus to generate images that share structural similarity with inputs.

\begin{figure}
  \centering
  \begin{subfigure}[t]{0.4\textwidth}
    \centering{}
    \includegraphics[scale = 0.75]{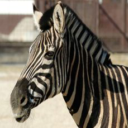}
    \caption{Real zebra image.}
  \end{subfigure}
  \begin{subfigure}[t]{0.4\textwidth}
    \centering
    \includegraphics[scale = 0.75]{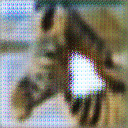}
    \caption{Generated horse image.}
  \end{subfigure}
  \caption{Generators quickly learn near-identity mapping at training epoch 3.}
  \label{fig:near-idt}
\end{figure}

\begin{figure}
  \centering
  \begin{subfigure}[t]{0.4\textwidth}
    \centering
    \includegraphics[scale = 0.75]{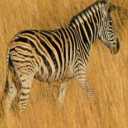}
    \caption{Real zebra image.}
  \end{subfigure}
  \begin{subfigure}[t]{0.4\textwidth}
    \centering
    \includegraphics[scale = 0.75]{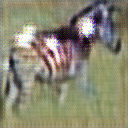}
    \caption{Generated horse image.}
  \end{subfigure}
  \caption{Generators quickly learn color mapping at training epoch 10.}
  \label{fig:color-map}
\end{figure}

\paragraph{Guide training} During experiments, we observe that the cycle consistency guides training by quickly driving generators to output images similar to inputs with simple color mappings. As shown in Figure~\ref{fig:near-idt}, the generator learns a near-identity mapping as early as training epoch 3 out of total 200. In Figure~\ref{fig:color-map}, we see that the generator learns to map yellow grass to green grass in zebra $\rightarrow$ horse translation at training epoch 10 out of 200. Upon inspecting the training dataset, we see that images from horse dataset generally have greener grass than those from zebra dataset. Because color mappings are often easily reversible, cycle consistency loss and GAN loss are particularly good at jointly guiding generators to output images with correct colors, which are crucial to whether they look realistic.

\paragraph{Regularize} Cycle consistency can also be viewed as a form of regularization. By enforcing cycle consistency, CycleGAN framework prevents generators from excessive hallucinations and mode collapse, both of which will cause unnecessary loss of information and thus increase in cycle consistency loss.

\afterpage{%
  \begin{figure}
    \centering
    \includegraphics[scale = 0.4]{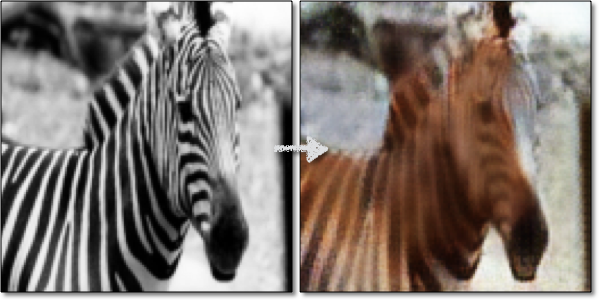}
    \caption{Unrealistic texture due to cycle consistency.}
    \label{fig:unrealistic-texture}
  \end{figure}
}

\afterpage{%
  \begin{figure}
    \centering
    \includegraphics[scale = 0.4]{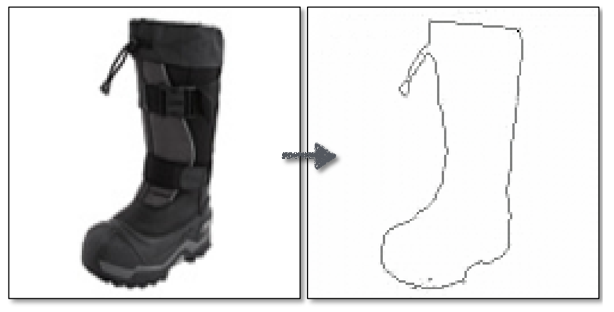}
    \caption{Generator must encoded color information in edges due to cycle consistency.\protect\footnotemark}
    \label{fig:unrealistic-encoding}
  \end{figure}
  \footnotetext{Figure adapted from \cite{zhu2017unpaired}.}
}

\paragraph{Unrealistic artifacts} Great as it is, cycle consistency is not without issues. Cycle consistency is enforced at pixel level. It assumes a one-to-one mapping between the two image domains and no information loss during translation even when loss is necessary. Consider the zebra $\rightarrow$ horse translation shown in Figure~\ref{fig:unrealistic-texture}, the generator cannot completely remove zebra texture because of cycle consistency. In the shoe $\rightarrow$ edges translation shown in Figure~\ref{fig:unrealistic-encoding}, also due to cycle consistency, color of the boot must be (potentially unperceptibly) somehow encoded in the result edges image, which causes unwanted artifacts.

\section{Better cycle consistency}
Cycle consistency is great. But as shown above in Section~\ref{sec:cycle}, it is sometimes too strong an assumption and causes undesired results. In this section, we propose three changes to cycle consistency aiming to solve the aforementioned issues.

\subsection{Cycle consistency on discriminator CNN feature level} Information is almost always lost in the translation process. Instead of expecting CycleGAN to recover the original exact image pixels, we should better only require that it recover the general structures. For example, in the zebra-to-horse translation, as long as the reconstructed zebra image has realistic zebra stripes, be it horizontal or vertical, identical to original ones or not, the cycle should be considered consistent. We enforce this weaker notion of cycle consistency by including an $L1$ loss on the CNN features extracted by the corresponding discriminator, which hopefully has learned good features on the image domain we are interested in. Specifically, the modified cycle consistency loss for one direction is now defined as a linear combination of CNN feature level and pixel level consistency losses:
\begin{equation}
  \tilde{\mathcal{L}}_\text{cyc}(G, F, D_X, X, \gamma) = \mathbb{E}_{x \sim p_\text{data}(x)}[\gamma \lVert f_{D_X}(F(G(x))) - f_{D_X}(x) \rVert_1 + (1 - \gamma) \lVert F(G(x)) - x \rVert_1], \label{eq:cnn-cc}
\end{equation} where $f_{D_{(\cdot)}}$ is the feature extractor using last layer of $D_{(\cdot)}$, and $\gamma \in [0, 1]$ indicates the ratio between discriminator CNN feature level and pixel level loss. This approach is similar to the deep perceptual similarities metric (DeePSiM) in GAN setting introduced in \cite{dosovitskiy2016generating}. DeePSiM also uses a combination of pixel level distance and CNN feature level distance, where the CNN can be fixed, such as VGGNet, or trained, such as generator or discriminator.

In practice, we observe that during training, it is best that $\gamma$ vary with epoch. In particular, $\gamma$ should start low because discriminator features are not good at beginning, and gradually linearly increase to a high value close but not equal to $1$ because some fraction of pixel level consistency is needed to prevent excessive hallucination on background and unrelated objects in the images.

\subsection{Cycle consistency weight decay}
As shown in Section~\ref{sec:cycle}, cycle consistency loss helps stabilizing training a lot in early stages but becomes an obstacle towards realistic images in later stages. We propose to gradually decay the weight of cycle consistency loss $\lambda$ as training progress. However, we should still make sure that $\lambda$ is not decayed to $0$ so that generators won't become unconstrained and go completely wild.

\subsection{Weight cycle consistency by quality of generated image}

\begin{figure}
  \centering
  \begin{subfigure}[t]{0.32\textwidth}
    \centering
    \includegraphics[scale = 0.75]{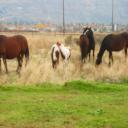}
    \caption{Real horse image.}
  \end{subfigure}
  \hfill
  \begin{subfigure}[t]{0.32\textwidth}
    \centering
    \includegraphics[scale = 0.75]{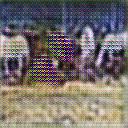}
    \caption{Generated zebra image.}
  \end{subfigure}
  \hfill
  \begin{subfigure}[t]{0.32\textwidth}
    \centering
    \includegraphics[scale = 0.75]{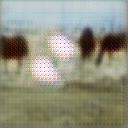}
    \caption{Reconstructed horse image.}
  \end{subfigure}
  \\
  \begin{subfigure}[t]{0.32\textwidth}
    \centering
    \includegraphics[scale = 0.75]{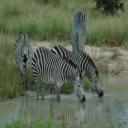}
    \caption{Real horse image.}
  \end{subfigure}
  \hfill
  \begin{subfigure}[t]{0.32\textwidth}
    \centering
    \includegraphics[scale = 0.75]{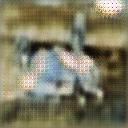}
    \caption{Generated zebra image.}
  \end{subfigure}
  \hfill
  \begin{subfigure}[t]{0.32\textwidth}
    \centering
    \includegraphics[scale = 0.75]{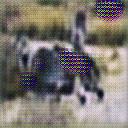}
    \caption{Reconstructed horse image.}
  \end{subfigure}
  \caption{Color inversion effect observed at training epoch 6.}
  \label{fig:color-inv}
\end{figure}

Sometimes early in training, we observe cases where generated image is very unrealistic and cycle consistency doesn't even make sense. For instance, in Figure~\ref{fig:color-inv}, the two generators, instead of trying to generate realistic images, learns color inversion mapping so that they can collectively decrease cycle consistency loss. In fact, once stuck in such local modes, the generators are unlikely to escape due to the cycle consistency loss. Therefore, enforcing cycle consistency on cycles where generated images are not realistic actually hinders training. To solve this issue, we propose to weight cycle consistency loss by the quality of generated images, which we obtain using the discriminators' outputs. Adding this change to Equation~\ref{eq:cnn-cc}, we have the new cycle consistency loss: \begin{equation}
    \begin{multlined}
      \tilde{\mathcal{L}}_\text{cyc}(G, F, D_X, X, \gamma) = \\
      \mathbb{E}_{x \sim p_\text{data}(x)}\left[D_X(x) \Big( \gamma \lVert f_{D_X}(F(G(x))) - f_{D_X}(x) \rVert_1 + (1 - \gamma) \lVert F(G(x)) - x \rVert_1 \Big)\right]
    \end{multlined}
\end{equation}

In particular, such a change dynamically balances GAN loss and cycle consistency loss early in training. It essentially urges the generators to first focus on outputting realistic image and to worry about cycle consistency later. 

It is worthwhile to note that gradient of this loss should not be backward propagated to $D_{(\cdot)}$ because cycle consistency is a constraint only on generators.

\subsection{Full objective}
Putting the above proposed changes together, the full objective at epoch $t$ it:

\begin{align}
  \mathcal{L}(G, F, D_X, D_Y, t) & = \mathcal{L}_\text{GAN}(G, D_Y, X, Y) + \mathcal{L}_\text{GAN}(F, D_X, Y, X) \\
  & + \lambda_t \tilde{\mathcal{L}}_\text{cyc}(G, F, D_X, X, \gamma_t) + \lambda_t \tilde{\mathcal{L}}_\text{cyc}(F, G, D_Y, Y, \gamma_t),
\end{align} where we suggest $\lambda_t$ to linearly decrease to a small value and $\gamma_t$ to linearly increase to a value close to $1$.

\section{Experiments}
We compare the proposed approach with original CycleGAN on \texttt{horse2zebra} dataset. In both experiments, we train with constant learning rate $0.0002$ for $100$ iterations and linearly decaying learning rate to $0$ for another $100$ iterations.

\begin{figure}
  \centering
  \includegraphics[scale = 0.3]{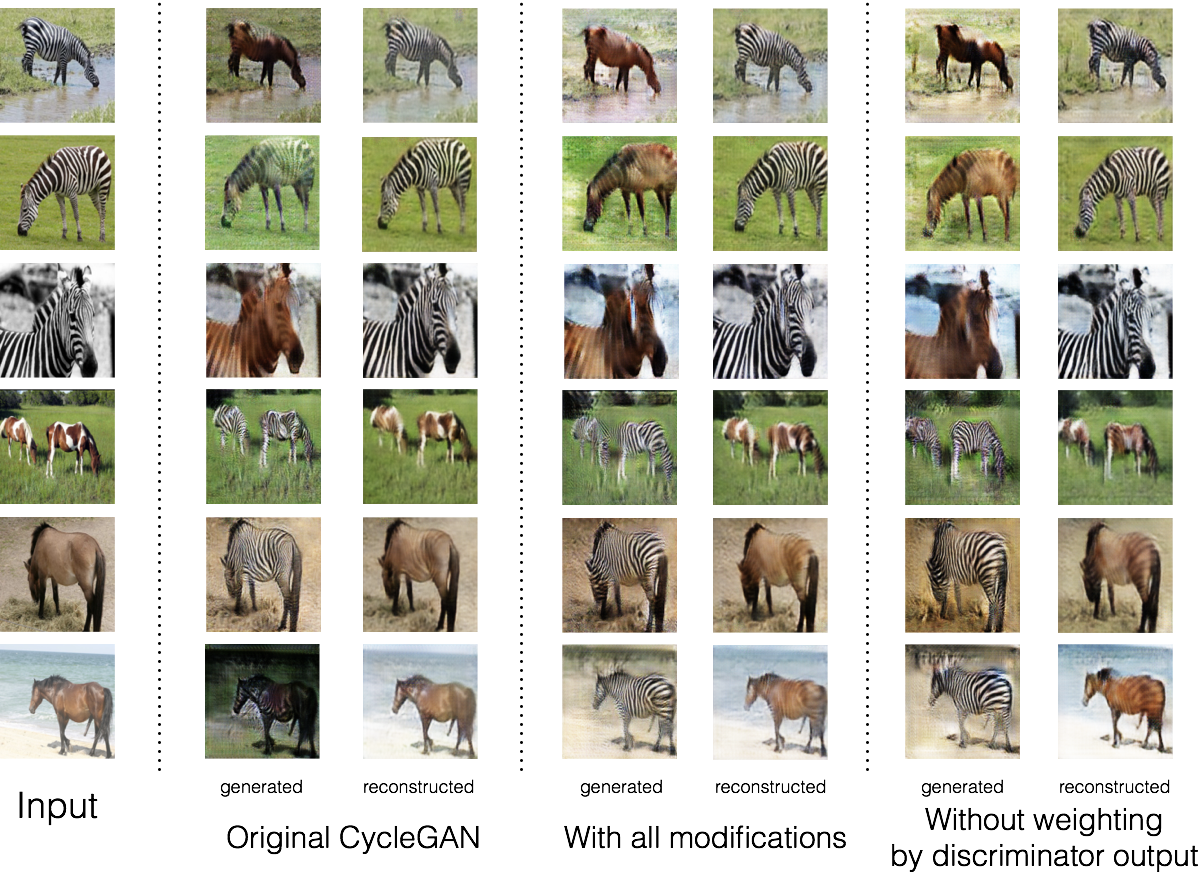}
  \caption{Comparison among original CycleGAN, CycleGAN with proposed modifications, and CycleGAN with proposed modifications except weighting cycle consistency by discriminator output on \texttt{horse2zebra} dataset. These images are hand picked from training set.}
  \label{fig:exp}
\end{figure}

Figure~\ref{fig:exp} shows the comparison on training set among original CycleGAN, CycleGAN with proposed modifications, and CycleGAN with proposed modifications except weighting cycle consistency by discriminator output.

As we can see, our proposed changes achieve better results with less artifacts than original CycleGAN. Specifically, although the reconstructed images are not as close to the original inputs, the generated outputs generally look more realistic. However, weighting cycle consistency by discriminator output doesn't contribute much to the result quality. We suspect that this is due to that discriminators are jointly trained with generators. During entire training, the discriminator outputs mostly stays around a constant value, which we observe to be about $0.3$. Therefore, we believe that using pretrained discriminators will make this modification actually have positive effect. In Section~\ref{sec:future-work}, we will discuss the potential approach of pretraining and fine-tuning discriminators in greater depth.

\begin{figure}
  \centering
  \includegraphics[scale = 0.3]{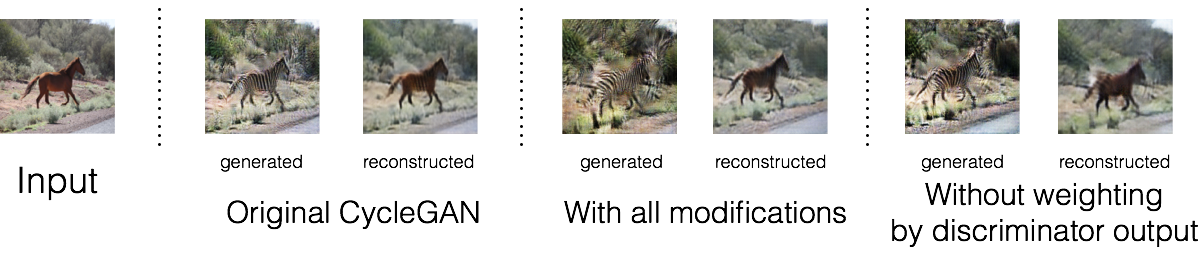}
  \caption{Failure case on \texttt{horse2zebra} dataset. }
  \label{fig:exp-fail}
\end{figure}

Nonetheless, we still found some cases where our modifications relax the cycle consistency may be too much so that it allows the generators to hallucinate unwanted artifacts, such as zebra textures around the horse in Figure~\ref{fig:exp-fail}. We think better parameter tuning will alleviate such issue.

\section{Future work}
\label{sec:future-work}
While our proposed approach achieves better results, there are still many exciting directions for us to explore, for example, how to tune the parameters, how to solve the one-to-many mapping problem. In this section, we describe several directions that we should investigate in future.

\paragraph{Parameter tuning} With the proposed changes, training CycleGAN now has many more parameters to tune, e.g., when and how to change $\lambda_t$ and $\gamma_t$, which discriminator to use as feature extractor, etc. During experiments, we found that the result quality is very sensitive to different parameters. Due to time limit, we are only able to try a few combinations. Experimenting for better parameters is definitely an important and future work direction.

\paragraph{Pretrain and fine-tune discriminators} Discriminators play an important role in two of three our proposed changes. However, since the discriminators are trained together with generators, they don't offer much in early stages. We believe that pretrained discriminators, either trained with CycleGAN task or initialized with pretrained CNN weights like AlexNet, along with fine-tuning, should give a considerable improvement over our current results. Moreover, because CycleGAN uses least-squares GAN, we in theory can over-train the discriminators and need not to worry about the diminishing gradients problem \cite{mao2016least}.

\paragraph{One-to-many mapping with stochastic input} Another exciting direction is to feed stochastic input to the generators so that they are essentially one-to-many mappings. However, it remains unknown how to include stochastic input into the architecture while still properly enforcing cycle consistency or some other training guidance and regularization. We attempted adding a noise channel to input and modifying generators to output an extra channel of data, which is encouraged to have same distribution as the input noise channel with the help of discriminators. However, we weren't able to achieve comparable results with the original CycleGAN within same period of training, likely due to a full channel of noise being too much randomness.

\paragraph{Generators with latent variables} We can consider the two image domains in CycleGAN translation task as sharing a latent space, and each generator as first mapping to latent space and then mapping to target domain. To incorporate this idea into CycleGAN framework, we can pick a certain layer in generator architecture as outputting latent variable, and think of the two generators as a pair of ``mutual encoders/decoders'' in the sense that encoding and decoding between a certain image domain and latent space are done in two different generators. Then, we can potentially enforce other notions of consistency, such as latent variable consistency and shorter cycle consistency ($\text{image domain} \rightarrow \text{latent space} \rightarrow \text{image domain}$).

\paragraph{Single discriminator for both directions} Since the two discriminators do classification on different domains, we can potentially replace them with one network that classifies among three classes, two image domains and fake images. Therefore, the discriminator sees data from both domains. Because the generators almost always are near-identity mapping at early stage of training, such a discriminator may better drive the generators towards right directions.

\section{Conclusion}
In this project, we identify and analyze several issues in CycleGAN framework caused by the cycle consistency assumption. In order to solve these issues, we propose changes to cycle consistency: adding $L1$ loss on the CNN features extracted by the corresponding discriminator, decaying the weight of cycle consistency loss as training progresses, and weighting cycle consistency loss by the quality of generated images. Training on the \texttt{horse2zebra} dataset, We show that experiment results on \texttt{horse2zebra} improve obviously. Last but not least, we point out several exciting future work directions to investigate.

\bibliographystyle{plainnat}
\bibliography{ref}

\end{document}